\documentclass[letterpaper, 10 pt, conference]{ieeeconf}  % Comment this line out if you need a4paper
\pdfoutput=1

\IEEEoverridecommandlockouts                              % This command is only needed if 
\usepackage{times}
\usepackage{epsfig}
\usepackage{graphicx}
\usepackage{amsmath}
\usepackage{amssymb}
\usepackage{subcaption}
\usepackage{algorithm}
\usepackage[noend]{algpseudocode}
\usepackage{balance}
\usepackage{color}
\usepackage{url}
\usepackage{authblk}
\makeatletter
\def\BState{\State\hskip-\ALG@thistlm}
\makeatother
\algdef{SE}[DOWHILE]{Do}{doWhile}{\algorithmicdo}[1]{\algorithmicwhile\ #1}

\graphicspath{ {figs/} }

\title{\LARGE \bf Towards Unsupervised Weed Scouting for Agricultural Robotics}

\author{David Hall$^{1}$\thanks{$^1$ The authors are with the School of Electrical Engineering and Computer
Science, Queensland University of Technology (QUT), Brisbane, Australia.
email: \{d20.hall, jason.kulk, c.mccool\}@qut.edu.au} \and Feras Dayoub$^{2}$\thanks{$^{2}$ The author is with the ARC Centre of Excellence for Robotic Vision, Queensland University of Technology (QUT), Brisbane, Australia.
http://www.roboticvision.org/ email: feras.dayoub@qut.edu.au} \and Jason Kulk$^{1}$ \and Chris McCool$^{1}$}

\begin{document}

\bstctlcite{IEEEexample:BSTcontrol}

\maketitle
\thispagestyle{empty}
\pagestyle{empty}

%%%%%%%%%%%%%%%%%%%%%%%%%%%%%%%%%%%%%%%%%%%%%%%%%%%%%%%%%%%%%%%%%%%%%%%%%%%%%%%%
\begin{abstract}

Weed scouting is an important part of modern integrated weed management but can be time consuming and sparse when performed manually.
%Automated weed scouting has been limited for widespread application partially due to a lack of adaptability in current state-of-the-art plant classification algorithms.
Automated weed scouting and weed destruction has typically been performed using classification systems able to classify a set group of species known \textit{a priori}.
This greatly limits deployability as classification systems must be retrained for any field with a different set of weed species present within them.
In order to overcome this limitation, this paper works towards developing a clustering approach to weed scouting which can be utilized in any field without the need for prior species knowledge.
We demonstrate our system using challenging data collected in the field from an agricultural robotics platform.
We show that considerable improvements can be made by (i) learning low-dimensional (bottleneck) features using a deep convolutional neural network to represent plants in general and (ii) tying views of the same area (plant) together.
Deploying this algorithm on in-field data collected by AgBotII, we are able to successfully cluster cotton plants from grasses without prior knowledge or training for the specific plants in the field.

\end{abstract}

%%%%%%%%%%%%%%%%%%%%%%%%%%%%%%%%%%%%%%%%%%%%%%%%%%%%%%%%%%%%%%%%%%%%%%%%%%%%%%%%
\section{Introduction}
%\TODO{Intro - Setup the problem of weed scouting in agriculture and how it has been looked into ith agricultural robotics before. Describe the benefits that this can have both for short term economic gains and weed killing efficacy and in the long term research of tracking and monitoring weed populations. Mention how automated approaches can speed up the process without the same need for concentrated effort by trained experts. Mention how the current approach of classification is useful but makes many assumptions which make widespread distribution difficult and how more adaptability is needed. Mention the challenges that arise due to this,the variability of how plants looked and the unconstrained environments they can grow in, the variation in the number of species and quantity of samples for each present.Describe what I plan to do, clustering, multiview clustering to improve clustering accuracy using descriptors made for dealing with plant classification.}

Integrated weed management is an increasingly crucial element to modern farming practice~\cite{Charles2011}.
Due to the notable increase in herbicide resistance within many weed species in recent years~\cite{Gilbert2013}, it has become increasingly important that weeds are dealt with in an appropriate manner which reduces herbicide resistance in weed species populations and maximizes the effectiveness of weed management.
A key element to effective integrated weed management is knowing which weed species are present in the field and in what quantities~\cite{Charles2011}~\cite{grdc2016} which we shall refer to as weed scouting.

Weed scouting is most commonly done through a manual inspection of the field using sampling techniques to get an impression of the weed species distributions.
This is a laborious process which seems ideal for automation where a robot can instead scout the entire field, having humans dictate weed management policy based upon the robot's findings.
While there have been some efforts in recent years to create robots for automated weed control and/or weed scouting, there are still critical areas which require improvement before widespread application of these systems is possible.

%% % % % % %MOVE TO THE LIT REVIEW % % % % % % % % %
%Currently, automated weed scouting and destruction has fallen under two main categories, general weed density calculation and weed species classification\TODO{probably need a ref}.
%These either provide a coverage of all weeds such as in\TODO{ref} or perform a species level classification for each plant as demonstrated in \TODO{REF}.
%While effective at their given tasks, they are not easily applicable for deployment on all farms to allow for integrated weed management strategies to be implemented.
%General weed density calculation, while effective for herbicide savings, does not provide any species level information and is ill-suited for integrated weed management.
%While providing species information, weed classification systems require advance knowledge of which species are expected to be found in each field, restricting its ability to be deployed to a variety of farms, each which may have their own plant species requirements \TODO{rethink that last sentence}.

\begin{figure}[t]
	\centering
	\includegraphics[width=0.85\linewidth]{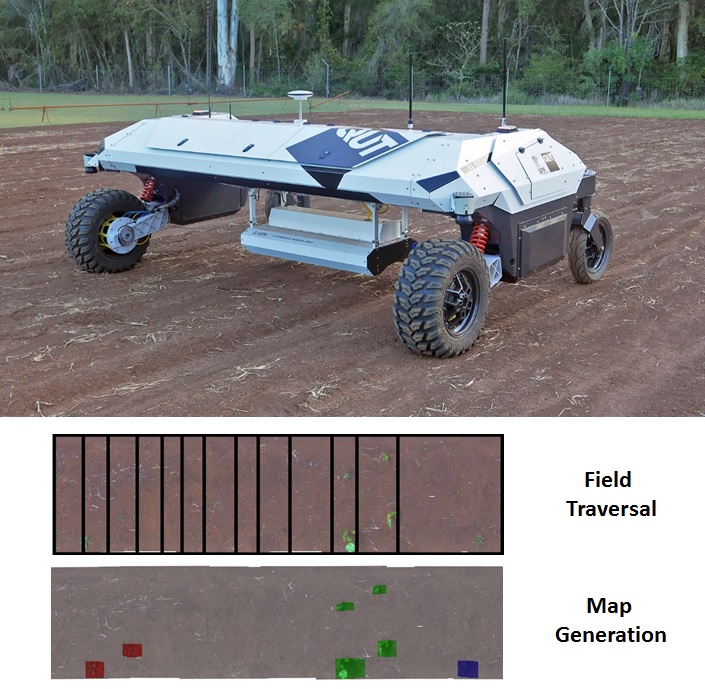}
	\caption{Simplified demonstration of weed scouting where the AgBotII robotic platform traverses the field and marks out weed locations. Each coloured box represents a different weed group. Field traversal image has undergone some image enhancement for visualization purposes. Best viewed in colour.}
	\label{fig:intro:coverImg}
\end{figure}

In this paper, we introduce a new unsupervised system for weed scouting for agricultural robotics utilizing weed clustering.
Clustering, unlike current state-of-the-art autonomous weed mapping systems, would be able to be applied in any field without requiring specific weed species knowledge in advance in order to generate weed distribution maps.
Weeds are clustered into visually similar groups which can then be identified by farmers without requiring manual inspection of each plant or a prior knowledge about the field.
We adapt current clustering algorithms to improve them for our task and demonstrate the first results of our clustering approach to weed scouting utilizing data collected from the AgBotII agricultural robotics platform (shown in Fig.~\ref{fig:intro:coverImg}) showing its applicability to real world conditions.

The contributions supplied in this paper are as follows.
We introduce a set of deep convolutional neural network (DCNN) bottleneck features trained on an unrelated plant dataset to be used within our unsupervised system for weed scouting, showing that these can improve upon results using normal DCNN features.
We introduce an image-locking system for hierarchical-type clustering algorithms which takes advantage of multiple tracked plant observations to improve clustering accuracy.
Finally, we introduce a new application-oriented evaluation metric.

\section{Literature Review}\label{sec:lit}
%Weed scouting is an important part of modern integrated weed management practice~\cite{Charles2011}.
%It allows for significant savings in herbicide use~\cite{gerhards2006practical} as well as the ability to perform long term analysis of the propagation of weed species within a single field and around the world, such as was done in~\cite{Pinke2016}.

Commonly, weed scouting has been done through a manual inspection of the field. 
However, and as shown in~\cite{Pedersen2006}, autonomous weed scouting can reduce costs by 20 \% over manual weed inspection.
%In~\cite{Pinke2016}, 262 fields were analysed manually using sample plots for each field over two years to examine different factors influencing weed species compositions in Hungarian soybean fields.
%This was effective for such a large study, but more detailed analysis for each field is limited.

In previous attempts at automated weed scouting and destruction, there have been two main approaches, either through treating all weeds as being the same~\cite{Sansao2012}, or through species classification~\cite{gerhards2006practical}.
The first approach, while effective for choosing herbicide dosages and examining where the highest infestations occur, is not particularly useful for integrated weed management strategies where each weed species must be treated in an appropriate manner.
The second approach uses supervised classification, which can achieve high accuracy, but requires \textit{a priori} knowledge of the weed species present in the field so that it can be trained to recognise them.
These issues restrict the widespread deployment of such approaches.

%One of the best practical examples of species identification being used to improve efficacy and herbicide distribution using an automated platform is found in~\cite{gerhards2006practical}.
%This had a classification accuracy of up to 86 \%, was capable of providing a species distribution density map and herbicide savings of up to 81 \%.
%One of the main weaknesses identified in classification systems, however is their ability to classify only a set number of known species which restricts their ability for widespread distribution and application.

Another approach which can address this problem is weed scouting based on unsupervised methods such as clustering. Clustering is the procedure of forming meaningful groups from samples~\cite{Grira2004}. In our case, the samples are images of weed plants in the field and the groups are the species. 

Clustering can be done with or without knowing the number of clusters ($K$). Clustering methods which need a known $K$ like k-means~\cite{MacQueen1967} are unrealistic for a scouting scenario where there can be no certainty about how many groups will be required. Therefore in this work, we focus on clustering algorithms which estimate $K$ as a part of the clustering process.

Of the techniques that do not require a known number of clusters, one of the most popular is hierarchical clustering which aims to iteratively merge or split clusters based on a distance metric between clusters in order to form cluster relationship trees~\cite{Johnson1967}. This allows for a more adaptable clustering system, with the number of clusters being calculated after the full trees have been computed, often through some stopping criterion being reached~\cite{Salvador2004}.
%Hierarchical clustering though has also been demonstrated to have weaknesses particularly with regards to propagating errors.
In~\cite{Frigui1999}, however, it was pointed out that hierarchical clustering is a static system where points once assigned to one cluster cannot be allocated to a new cluster, allowing for errors to be propagated through the system.

In order to overcome such problems, variations to hierarchical clustering have been developed. Competitive agglomerative clustering, demonstrated in~\cite{Frigui1997} and~\cite{Frigui1999}, does not merge clusters in the same sense as is done in hierarchical clustering.
Instead, clusters competes over the samples available and are slowly removed as their samples are taken by larger clusters.
This is similar to what is done within infinite Gaussian mixture models (IGMMs) and Dirichlet process mixture models (DPMMs) which iteratively go through a process of reassigning points to existing clusters and removes clusters through this process~\cite{Steinberg2010}.

Another approach which is able to perform clustering without knowing the number of clusters is affinity propagation~\cite{Frey2006,Dueck2007}. This powerful technique generates exemplars by iteratively updating how likely the each sample can serve as an exemplar and how likely each sample could belong to each point as an exemplar.
Through this process, eventually a consensus occurs and clusters are assigned to all samples. 

An approach which is more akin to traditional hierarchical clustering but which still takes measures to allow for reassigning data points is the hierarchical approach utilized in diarization~\cite{Wooters2008}. Diarization systems are used heavily in signal processing and attempt to answer the question ``who spoke when?''~\cite{AngueraMiro2012}.
They comprise of two elements, the segmentation of audio data into speaker and non-speaker segments and clustering speaker segments into specific speaker groups with each group corresponding to one individual. The clustering used in diarization differs from hierarchical clustering in that after merging two clusters, data is resegmented and the clusters retrained~.

Another important aspect of weed scouting based on clustering is image features. In recent years the development of features learnt using deep convolutional neural networks~\cite{Donahue2013} have proved beneficial in many recognition tasks.
Deep neural network architectures such as those demonstrated in~\cite{Krizhevsky2012} and~\cite{Szegedy2015} can be fine-tuned as described in~\cite{Donahue2013} in order to obtain features more suited to the precise recognition task to be performed.

%This could potentially aid in clustering, having features trained for separating the types of objects desired, however these features are typically very high dimensional features, a trait which is typically undesirable for clustering tasks and must be addressed through techniques such as generating bottleneck features~\cite{ge2015content}.

As shown above, current automated weed scouting methods are based on supervised learning methods which need \textit{a priori} knowledge about the weeds that exist in the field. This defies the purpose of the scouting process (i.e finding what weed plants exist in the field). This shortcoming makes unsupervised clustering methods an attractive solution and more suitable approach for the task.

\section{Methodology}\label{sec:method}
The algorithm used in our weed scouting system for agricultural robotics consists of four main stages. 
These are plant detection and segmentation, feature extraction, species clustering, and field mapping. 
The following sections explain these stages in more detail. %Our main contributions in this work are in the fields of feature extraction and species clustering. All parts of our weed scouting algorithm shall all be explained in the following subsections.

\subsection{Plant Detection and Segmentation}\label{sec:method:segment}
\begin{figure}[t]
	\centering
	\begin{subfigure}[b]{0.235\linewidth}
		\centering
		\includegraphics[width=\linewidth]{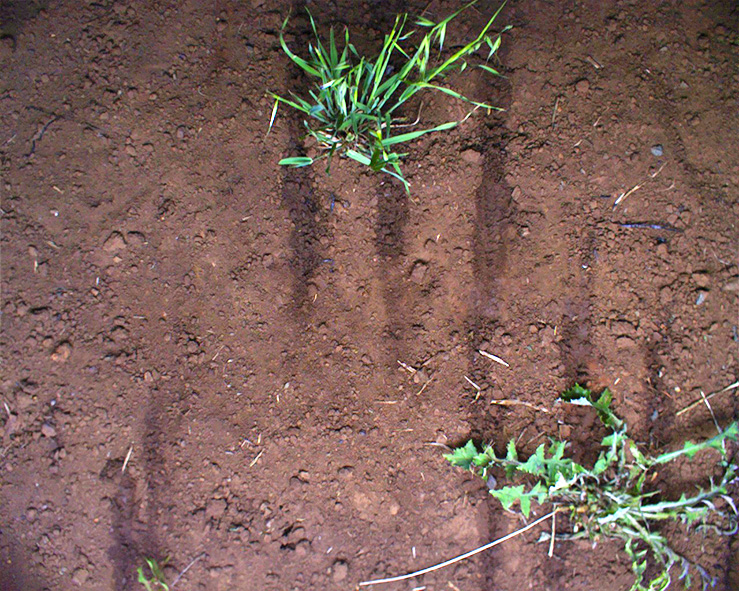}
		\caption{}
	\end{subfigure}
	\begin{subfigure}[b]{0.235\linewidth}
		\centering
		\includegraphics[width=\linewidth]{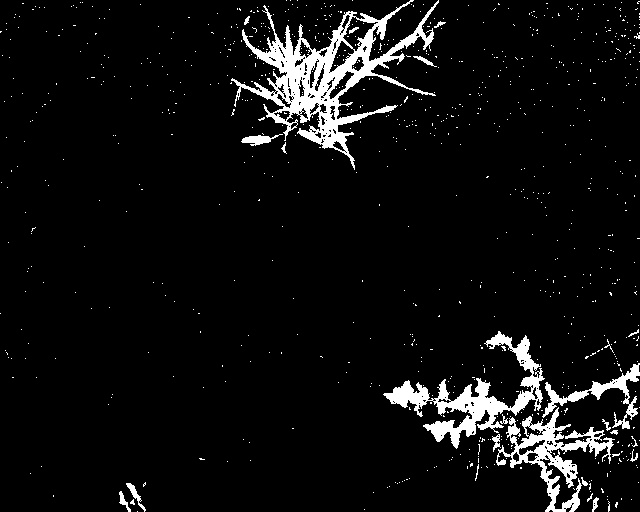}
		\caption{}
	\end{subfigure}
	\begin{subfigure}[b]{0.235\linewidth}
		\centering
		\includegraphics[width=\linewidth]{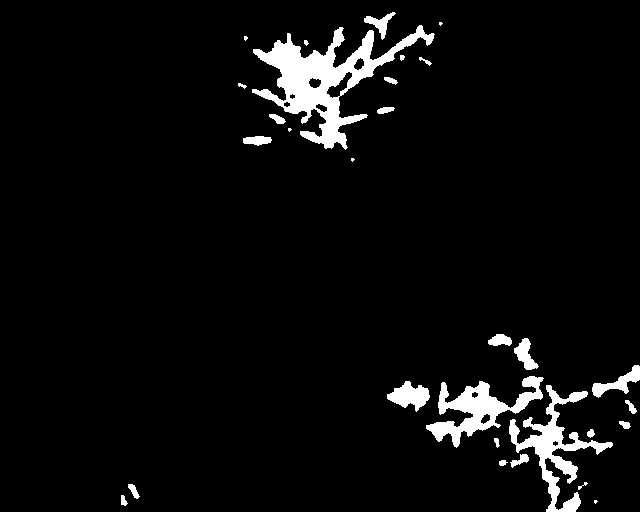}
		\caption{}
	\end{subfigure}
	\begin{subfigure}[b]{0.235\linewidth}
		\centering
		\includegraphics[width=\linewidth]{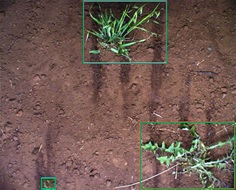}
		\caption{}
	\end{subfigure}
	\caption{From left to right (a) the original image, (b) the result of per-pixel segmentation after thresholding, (c) the filtered image after erosions and dilations (final mask), and (d) the detection regions, indicated by the bounding boxes.}
	\label{fig:method:detect}
\end{figure}

In our work, detection was performed using only colour imagery without any extra multi-spectral imaging. Inspired by~\cite{philipp2002improving}, our segmentation method utilizes multiple colour spaces for weed detection. 
%It is well known that the RGB colour space is extremely dependent upon illumination conditions, thus we only use colour-opponent colour spaces and cylindrical representations which provide some inherent robustness to illumination. 
%We utilize the following colour spaces: HSV, Lab, and Luv; in combination with each other, as they provide complementary representations of the same information. 
%A common factor for each of these colour spaces is that lightness (or brightness) is represented by one component and not present in all three components (as is the case for RGB).
%For this reason we discard this illumination information when we combine the colour spaces, forming our final colour descriptor used for segmentation. 
%Below we briefly list the properties of each of these colour spaces.
%%
The colour spaces utilized are summarized below.
\begin{description}
	\item[HSV] is a cylindrical colour space consisting of hue (H), saturation (S) and value (V), or brightness.
	\item[Luv] is a perceptually uniform colour space where lightness is captured by the component L.
	\item[Lab] is an opponent colour space where a and b are the colour-opponent dimensions.
\end{description}
Our plant detector uses a multivariate Gaussian trained with the following feature vector [H,S,u,v,a,b]. Model parameters and the threshold probability for the Gaussian are learnt on a training set of manually annotated images which are not related to the dataset used in our clustering experiments. The full detection pipeline is shown in Fig~\ref{fig:method:detect}. When applied to a new image, the Gaussian provides a per-pixel log-likelihood map, or segmentation map, similar to that in Figure~\ref{fig:method:detect} (b). As this is often noisy, to convert the segmentation map into weed regions we first remove noise from the image and then search for connected regions. Contiguous regions are found from this binary image by finding the contours~\cite{Suzuki85_1} and those which are close are merged to form a single weed segmentation mask which can be utilized later on in our clustering pipeline for feature extraction.

\subsection{Feature Extraction}\label{sec:method:feats}
Effective plant species clustering is highly dependant on the feature representation given for each plant. In our experiments we use and evaluate two main categories of feature, these being hand-crafted features (HCFs) typically used in plant classification for agricultural robotics and learnt deep convolutional neural network (DCNN) features.

We utilize the same HCFs as are used within~\cite{hall2015evaluation} and~\cite{Haug2014}, consisting of the following shape features: plant perimeter, plant area, length of plant mask skeleton, compactness, convexity, length of plant mask skeleton/plant perimeter; and the following statistical reflectance features: minimum, maximum, range, mean, median, standard deviation, kurtosis, and skewness of the plant pixel intensities.
Further detail of how each of these features are extracted can be found within~\cite{Haug2014}. As in~\cite{hall2015evaluation}, we analyse the use of two sets of HCFs, one containing all of the shape and reflectance features described, and a scale-robust subset which discludes the features which are affected by image scale (perimeter, area, and length of skeleton). These two feature sets will be referred to as hcf and hcf-scale-robust within this paper.

%Shape features require the use of a plant mask and so we utilize the the plant mask generated for each plant as described in section~\ref{sec:method:segment} for our pipeline.
%As we do not have access to NIR information as was used for calculating the reflectance features in~\cite{Haug2014}, we use the NExG grayscale vegetative index image as done in~\cite{hall2015evaluation}.
%Also, like~\cite{hall2015evaluation}, we analyse the use of two different sets of HCFs, one 15 dimensional descriptor containing all of the shape and reflectance features described, and one 12 dimensional scale-robust subset which discludes the features which are affected by image scale, these being perimeter, area, and length of skeleton.
%These two feature sets will be referred to as hcf and hcf-scale-robust within this paper.

%Due to the high success found in using discriminative deep convolutional neural network (DCNN) features for classification~\cite{krizhevsky2012imagenet}, 
We also used deep DCNN features for our clustering experiments. %We follow the procedure described in~\cite{Donahue2013} where features are first learnt on a big but general dataset such as ImageNet before adapting it to a more specific fine-grained domain such as birds~\cite{Donahue2013}.
Namely, the powerful Inception architecture~\cite{Szegedy2015} known as GoogLeNet which is a deep network with 22 layers. This is first pre-trained on ImageNet and fine-tuned on a training subset of the leaf images from the PlantCLEF dataset~\cite{Bonnet2015}.
%Features are extracted from the output of the average pooling layer of the GoogleNet model.

While learnt features are known to be highly descriptive, they come a cost of high dimensionality. It is known that a high dimension feature representation can cause a clustering system to suffer due to the ``curse of dimensionality''. In order to alleviate this issue, we follow the procedure shown in~\cite{ge2015content} to generate so called bottleneck features which are both discriminative and low dimensional. 
%We add a 128 neuron fully connected layer between the average pooling layer and the softmax layer of the GoogleNet architecture. 
%\TODO{CHECK THIS IS STILL THE CASE}
%Parameters for bottleneck feature are initialized with a zero mean and 0.001 standard deviation. 
This network is fine-tuned on the same training images from PlantCLEF as the original architecture. We shall refer to these as bottleneck features throughout this paper.
%As the networks described require a whole image to extract features from, in our pipeline we use the image within the bounding box region extracted by our detection algorithm as demonstrated in Section~\ref{sec:method:segment}.
%In order to ensure that the background is not considered as important information, which can occur for convnets, we segment out the background from this image using the plant mask.

\subsection{Clustering Algorithms}\label{sec:method:clust_algorithms}
We analyse several potential clustering algorithms that can be utilized within our weed scouting pipeline. %each of which does not require a known number of clusters, an important aspect required for an unsupervised weed scouting system as stated in~\ref{sec:lit}.
%We evaluate the variation of hierarchical clustering typically used in diarization (which we shall refer to as diarization clustering), hierarchical clustering, DPGMM clustering and affinity propagation as the baseline for our task.
The following sections shall describe these algorithms and how they are implemented.

\subsubsection{Diarization Clustering}\label{sec:method:diarize}
%Diarization as used in signal processing attempts to answer the ``who said what and when'' question.
The clustering methods used within diarization to group speakers together fall under two main categories, these being bottom-up or top-down diarization~\cite{AngueraMiro2012}.
In this work we utilize the clustering technique used in bottom-up diarization which is a variation of agglomerative hierarchical clustering~\cite{chen1998speaker} where clusters are iteratively merged together based upon a relative distance metric until an ending criterion is reached.%We shall refer to this as diarization clustering for the remainder of this paper.

In this work, we model the diarization clustering system heavily on the early diarization process described in~\cite{Zhou2000}. This earlier process was used as, unlike in later methods~\cite{Wooters2008}, it did not utilize a hidden Markov model which requires a link between consecutive observations which is not possible in our application.
We modify this early algorithm slightly however by, instead of using a single iteration of resegmentation and retraining, using multiple iterations until cluster stability is reached. Using this approach, clusters tend to be more stable before merging which leads to more accurate clusters being generated.
Our modified diarization clustering algorithm used within this work is shown in Algorithm~\ref{alg:method:diarize}.
In this bottom-up system, each cluster is modelled by a multivariate Gaussian.
We also add a further improvement specific to our application which we call cluster locking which will be explained further in Section~\ref{sec:method:diarize:locking}.

\begin{algorithm}[t]
	\begin{algorithmic}[1]
		\Procedure{Cluster}{}
		\State Initialize $n$ Clusters $\boldsymbol{\Omega}=\{\omega_1, \omega_2, ... \omega_n\}$
		\Do
		\While {$\textbf{not}$ stable($\boldsymbol{\Omega}$)}
		\State segmentClusters($\boldsymbol{\Omega}$)
		\State retrainClusters($\boldsymbol{\Omega}$)
		\EndWhile
		%		($\omega_a, \omega_b$) = $\min_{i,j}\text{KL2}(\omega_i, \omega_j)$
		\State $\omega_a, \omega_b = \text{closestPair}(\boldsymbol{\Omega}$)
		\State validMerge = evaluateMerge($\omega_a, \omega_b$)
		\If {validMerge}
		\State mergeClusters($\omega_a, \omega_b$)
		\EndIf
		\doWhile validMerge
		\State $\textbf{end}$
		\EndProcedure
	\end{algorithmic}
	\caption{Our diarization clustering algorithm consists of an over segmented cluster initialization (line 2) followed by a repeated loop of cluster resegmentation (line 5), cluster retraining (line 6), finding the closest clusters (line 7), determining if merging these clusters is beneficial (line 8) and if so, cluster merging the two clusters (line 10). Below $\boldsymbol{\Omega}$ is the collection of all current individual clusters $\omega$}
	\label{alg:method:diarize}
\end{algorithm}

%\begin{figure}[t]
%	\centering
%	\includegraphics[width=0.9\linewidth]{diarization_example}
%	\caption{Simplified illustrated example of optimization and merging steps within a diarization based clustering algorithm}
%	\label{fig:method:diarization:diarize_example}
%\end{figure}

We use symmetric Kullback-Leibler (KL2) distance~\cite{Zhou2000} to measure the distance between clusters and the Bayesian information criterion~\cite{Schwarz1978} as our stopping criterion.

\subsubsection{Cluster Locking}\label{sec:method:diarize:locking}
In our task, we aim to cluster each image of a plant into a group of similarly looking plants.
We can, however, utilize the fact that our robotic platform will be able to take multiple images of the same plant to enhance our clustering accuracy.
For instance, a robot may take several images of the same plant as it traverses over it, allowing for more information to be available for that one plant.
Rather than treat each image as its own plant, we can initialize plant images which are in approximately the same real-world location as being the same plant.
Using this we can generate initial clusters which have a better model representation of the single plant than each individual image would have.

In order to keep the high accuracy and improved plant models throughout clustering, we must also enforce that these images of the same plant, must be locked together throughout the optimization step of the diarization clustering procedure.
We do this by adding in an extra step checking after segmentation if samples of the same plant have been kept together or if they have been separated.
If the latter is true, then for each cluster containing the same plant sample points, we calculate the combined log-probability that each sample point belongs to that cluster.
The cluster which contains the highest log-probability is then given all samples of that plant.
This ensures that the samples of the plant are kept together while still allowing diarization to optimize the clusters after each iteration of cluster merging.
This leads to improved cluster models that better represent each of the plants contained within.
We refer to this new adaptation of diarization clustering as diarization-locked clustering throughout the remainder of this work.

\subsubsection{Hierarchical Clustering}\label{sec:method:hierarchy}
Agglomerative hierarchical clustering is a technique of progressively merging small clusters together based upon their relative distances to each other~\cite{Johnson1967}.
It merges the closest clusters iteratively to form a hierarchical clustering tree otherwise known as a dendrogram and chooses the number of clusters based on some predetermined stopping criterion.
Typically this begins with treating each data point as a cluster but can be given an initial over-segmentation for initialization.

In order to be consistent, in our work we utilize the same initialization in tests as is done for diarization and utilize the same distance metric and stopping criterion.
The only difference between the diarization clustering and hierarchical clustering used within this work is that diarization has the extra optimization step as explained in Section~\ref{sec:method:diarize}.

We also improve hierarchical clustering through using the same initialization as is done for the diarization-locked clustering algorithm described in Section~\ref{sec:method:diarize:locking} which we refer to as hierarchical-locked in our experiments.
While there is no actual locking procedure required throughout the clustering algorithm due to the lack of optimization in hierarchical clustering, the improved initialization which takes advantage of multiple observations of the same plant improves clustering accuracy and so is considered its own algorithm in our tests.

\subsubsection{Dirichlet Process Gaussian Mixture Model (DPGMM)}\label{sec:method:dpgmm}
In this work, one of the techniques we evaluate for use in the clustering step of the diarization pipeline is Dirichlet process Gaussian mixture model (DPGMM) clustering~\cite{Ferguson1973}.
In our work this is defined as a form of infinite Gaussian mixture model (IGMM)~\cite{Rasmussen1999}, which is a form of Gaussian mixture model (GMM) which does not require the number of Gaussian components to be known in advance, keeping with the core principle of what a diarization system should be capable of.

This system assumes an infinite number of clusters exist but that only a few are present in the given dataset, where each cluster is modelled using a Gaussian distribution.
The number of clusters chosen is dependant on an aggregation parameter $\alpha$.
When $\alpha$ is large, the number of clusters chosen is typically also large and the same applies for small values of $\alpha$.
Once the DPGMM is generated, the Gaussians within the model are used to formulate clusters with each data point being assigned to the cluster with the Gaussian which best represents them.
Further details about DPGMMs and IGMMs can be found within~\cite{Ferguson1973} and~\cite{Rasmussen1999}.
For our experiments, $\alpha$ was chosen experimentally for each test.

\subsubsection{Affinity Propagation (AP)}\label{sec:method:ap}
The final technique used in our clustering experiments is affinity propagation~\cite{Frey2006,Dueck2007}.
As with the other techniques used here, affinity propagation does not require a known number of clusters.
This technique generates exemplars by iteratively updating how likely each point can serve as an exemplar and how likely each point could belong to each point as an exemplar.
Through this process, eventually a consensus occurs and clusters are assigned to all points.
More details can be found within~\cite{Frey2006} and~\cite{Dueck2007}.

\subsection{Mapping}
To visualise the performance of the species clustering in the field a map was constructed from the image data set collected of the field. 
Each of the images in the data set were stitched together to form a large continuous mosaic of the field. 
The images from the data set are initially placed on to the mosaic using the position and rotation from the GPS and INS. 
To compensate for the inaccuacies in the GPS and INS values we match features extracted using~\cite{Rublee2011} between each pair of adjacent images. 
Based on the median of the difference in poses of the features we apply an affine transformation to align the overlapping portions of the images. The output of the clustering was then overlayed on to the map to display both the spatial accuracy and the clustered plant species.
\section{Experimental Setup}\label{sec:setup}
In order to test our unsupervised scouting for agricultural robotics system and evaluate the effects of the locking described in Section~\ref{sec:method:diarize:locking} we collected evaluation data using a robotics platform was required and developed a new method for evaluating the effectiveness of the clustering achieved which was appropriate for our application.
The data collection and evaluation metric used in our experiment are explained in the following two subsections.

\subsection{Evaluation Data Collection}\label{sec:setup:dataset}
\begin{figure}[t]
	\centering
	\setlength{\fboxsep}{0pt}
	\begin{subfigure}[b]{0.235\linewidth}
		\centering
		\fbox{\includegraphics[width=\linewidth]{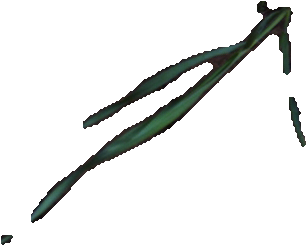}}
		\caption{Wild Oats}
	\end{subfigure}
	\begin{subfigure}[b]{0.235\linewidth}
		\centering
		\fbox{\includegraphics[width=\linewidth]{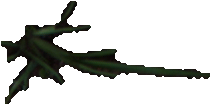}}
		\caption{Feathertop}
	\end{subfigure}
	\begin{subfigure}[b]{0.235\linewidth}
		\centering
		\fbox{\includegraphics[width=\linewidth]{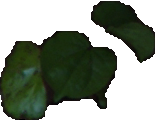}}
		\caption{Cotton}
	\end{subfigure}
	\begin{subfigure}[b]{0.235\linewidth}
		\centering
		\fbox{\includegraphics[width=\linewidth]{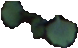}}
		\caption{Sowthistle}
	\end{subfigure}
	\caption{Example of each plant species within our dataset}
	\label{fig:setup:data_examples}
\end{figure}

Our tests for unsupervised weed scouting were performed using data collected from the Agbot II agricultural robotics platform.
Data was collected from sections of a single field which was seeded with four specific weed species, two grasses and two broad-leaves.
The species grown were wild oats, feathertop, cotton and sowthistle as shown in~\ref{fig:setup:data_examples}.
A summary of the data collected can be seen in Table~\ref{tbl:setup:dataset}.
It should be noted that evaluation is done on the image level and the number of plants is given merely as reference.
Plant-level observations are used only for diarization-locked and hierarchical-locked clustering methods as described in Section~\ref{sec:method:diarize:locking}.

\begin{table}[t]
	\centering
	\caption{Summary of test dataset}
	\label{tbl:setup:dataset}
	\begin{tabular}{|c|c|c|c|c|}
		\hline
		& \textbf{Wild Oats} & \textbf{Feathertop} & \textbf{Cotton} & \textbf{Sowthistle} \\ \hline
		\textbf{\# Images} & 113                & 90                  & 157             & 11                  \\ \hline
		\textbf{\# Plants} & 33                 & 27                  & 49              & 3                   \\ \hline
	\end{tabular}
\end{table}

The process for dataset compilation was as follows.
Image frames were extracted at a rate of 5 Hz as the AgBotII traversed the desired sections of the field.
%No covering surrounded the viewed area to protect from the sun, but a strobe light was used to artificially illuminate the area to a desirable level. 
Plant regions were extracted from the image frames into cropped individual images with segmented backgrounds as shown in Fig.~\ref{fig:setup:data_examples} using the segmentation algorithm described in Section~\ref{sec:method:segment}.
These cropped images were manually labelled according to the four species described above.

The shape HCFs used in this work as described in Section~\ref{sec:method:feats} were extracted using the largest contour of the segmentation mask for each image and as our images only contained RGB information, the statistical features were calculated using the normalized excessive green vegetative index image as was done in~\cite{hall2015evaluation}.
As explained in Section~\ref{sec:method:feats}, we also extract two forms of DCNN features.
GoogLeNet features are extracted from the output of the average pooling layer of the fine-tuned GoogLeNet model.
To generate bottleneck features we added a 128 neuron fully connected layer between the average pooling layer and the softmax layer of the GoogLeNet architecture before subjecting the network to the same fine-tuning as was done for the GoogLeNet model.
Features were then extracted from this layer.
We also generate a series of features where the hand-crafted and DCNN features are concatenated together.
All features undergo L$_2$ normalization before being fed into any of the clustering algorithms.
For concatenated features, L$_2$ normalization is performed on the separate feature types before merging and no further normalization is performed after this.

As the robot traversed the field, some plants were detected multiple times and each instance is included in the final dataset.
In order to implement the improved initialization and image locking mentioned in Section~\ref{sec:method:diarize:locking}, cropped plant images were also manually grouped together if they were all from the same plant.
Clusters therefore could be initialized with data from all images which are associated with a single plant.
While this was done manually for our dataset generation, a similar result could be achieved for an online robotic application using relative GPS position of each extracted plant region such as is done on the BoniRob platform~\cite{Ruckelshausen2009}.
In all other instances, initialization treated each image as its own cluster, except for DPGMM which was initialized with 300 clusters.

\subsection{Evaluation Metric}\label{sec:setup:evaluation}
In order to quantifiably evaluate and compare the different clustering techniques examined within this work, an appropriate evaluation metric was required.
Typically in clustering analysis with a set goal in mind such as ours, clustering analysis is performed using pairwise variations of classic evaluation measures such as pairwise precision, recall, accuracy, and F1 score which are described in more detail in~\cite{Manning2008}.

While effective, these metrics do not prioritize the goals of this project.
Our goal is to generate very few, pure clusters which can be examined by a farmer and then manually classified by them.
This therefore requires a high priority on having very pure clusters which are separated into as few groups as possible so as to save the farmer time (i.e. 20 very pure clusters of the same species is not very desirable).
Here we define the purity of cluster $k$ in terms of class $c$ as follows:

\begin{equation}
Purity(k|c) = \frac{n_k^c}{n_k}
\end{equation}

\noindent where $n_k^c$ is the number of samples held in cluster $k$ of class $c$, $n_k$ is the total number of samples within cluster $k$,

\begin{figure}[t]
	\centering
	\includegraphics[width=\linewidth]{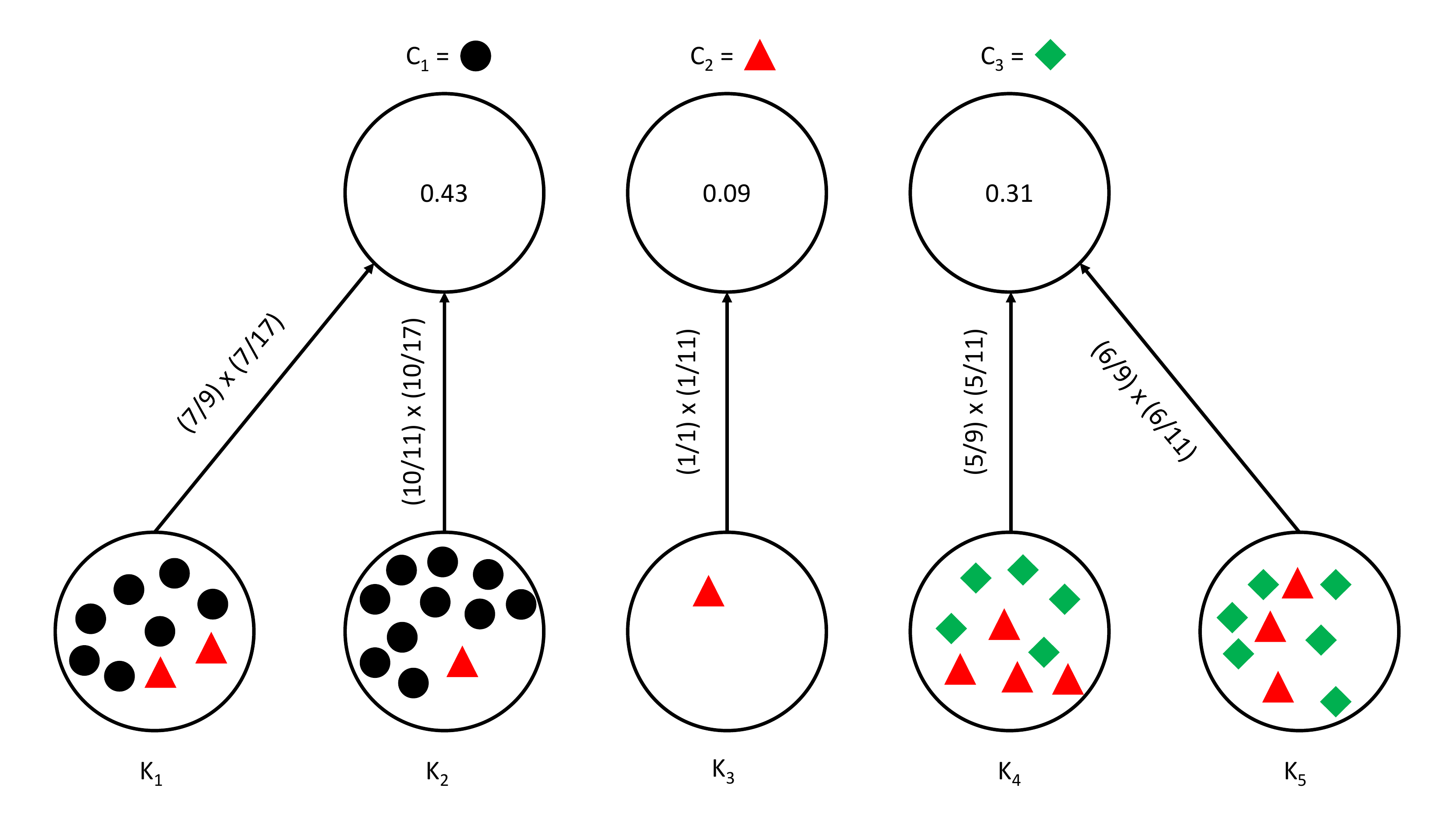}
	\caption{Simplistic scenario to describe the calculation of the DScore. Each cluster is assigned to a class, as indicated by the arrows, based on which class is dominant within the cluster. The score for each class is the sum of its assigned cluster's purities, weighted by the proportion of the class' points owned by said cluster, divided by the number of clusters assigned to the class. For example, $K_1$ has a purity of $7/9$ as there are 7 samples of the dominant class $C_1$ (black dots) out of the total of 9 samples within cluster $K_1$. The weighted importance of $K_1$ towards the score for class $C_1$ is $7/17$ as it contains 7 of the total 17 $C_1$ samples. Weighted addition of $K_1$ and $K_2$s contributions is then divided by the number of clusters assigned to $C_1$ which is 2 to give $Score(C_1) = ((7/9\times7/17) + (10/11\times10/17))/2 = 0.43$. The score for each class is shown in the class circle. The final DScore for the clustering above is 0.28}
	\label{fig:setup:metrics:DScore}
\end{figure}

In order to better analyse our results in the terms which are important to our application, a new metric was generated which we shall refer to as the DScore.
First, each cluster is assigned to the class which dominates the cluster.
For example in Fig~\ref{fig:setup:metrics:DScore}, the first cluster ($K_{1}$) mostly contains black dots which are data points from class 1 ($C_{1}$) and so $K_{1}$ has been assigned to $C_{1}$.
In context this could mean that the $K_{1}$ has been assigned to the cotton class. 
We calculate an individual score for each class which is a weighted sum of cluster purities for all clusters divided by the number of clusters which the class has been separated into.
This division penalizes oversegmentation while the weighted purity prioritizes the importance of as many points as possible being part of pure clusters.
DScore is formulated as follows:

\begin{equation}
Score(c) = \frac{\sum_{k \epsilon G}Purity(k|c) \times  \frac{n_k^c}{n^c}}{N_K^c} 
\end{equation}
\noindent where $G$ is the set of clusters which have been assigned to class $c$, $n^c$ is the total number of samples of class $c$, and $N_K^c$ is the number of clusters assigned to class $c$.

The final DScore is then calculated as the average of the scores across all classes.

\begin{equation}
DScore = \frac{\sum_{i=1}^{N_C} score(i)}{N_C}
\end{equation}

\noindent where $N_C$ is the total number of classes.
This averaging ensures that each class is considered equally important to the final score, avoiding issues with unbalanced data.
If this was not done, the metric may presume very good clustering has been achieved despite one class being very poorly clustered such as the red triangles of $C_2$ in Fig~\ref{fig:setup:metrics:DScore}.

The DScore has a maximum value of 1 associated with perfect clustering and is a good measure for the effectiveness of clustering within the task of unsupervised weed scouting for agricultural robotics.
A full example is shown in Fig~\ref{fig:setup:metrics:DScore}.
\section{Results}\label{sec:results}
To demonstrate the effectiveness of our features and clustering techniques, we performed a thorough analysis across all feature combinations and clustering techniques described in Sections~\ref{sec:method:feats} and \ref{sec:method:clust_algorithms} respectively.
This includes eight different combinations of features and six clustering algorithms including our diarization-locked and hierarchical-locked clustering algorithms.
Analysis was performed both quantitatively and qualitatively as shall be demonstrated in the following subsections.

The segmentation algorithm, described in Section~\ref{sec:method:segment}, was derived using independent training and evalution data. 
On this independent data, it achieved an F1 score of 0.97 on the evaluation data which demonstrates the effectiveness of the approach.

\subsection{Clustering Quantitative Analysis}\label{sec:results:quant}
All quantitative analysis of the various algorithms and features was achieved using our new evaluation metric known as the DScore, described in full in Section~\ref{sec:setup:evaluation}.

\begin{table}[]
	\centering
	\caption{Best Clustering Result Comparison}
	\label{tbl:results:clust_compare}
	\begin{tabular}{|c|c|}
		\hline
		\textbf{Clustering Algorithm} & \textbf{DScore} \\ \hline
		AP                            & 0.16            \\ \hline
		DPGMM                         & 0.37            \\ \hline
		Diarization                   & 0.37            \\ \hline
		Diarization-locked   		  & 0.40            \\ \hline
		Hierarchical                  & 0.41            \\ \hline
		\textbf{Hierarchical-locked}  & \textbf{0.44}   \\ \hline
	\end{tabular}
\end{table}

\begin{figure*}[t]
	\centering
	\includegraphics[width=0.7\linewidth]{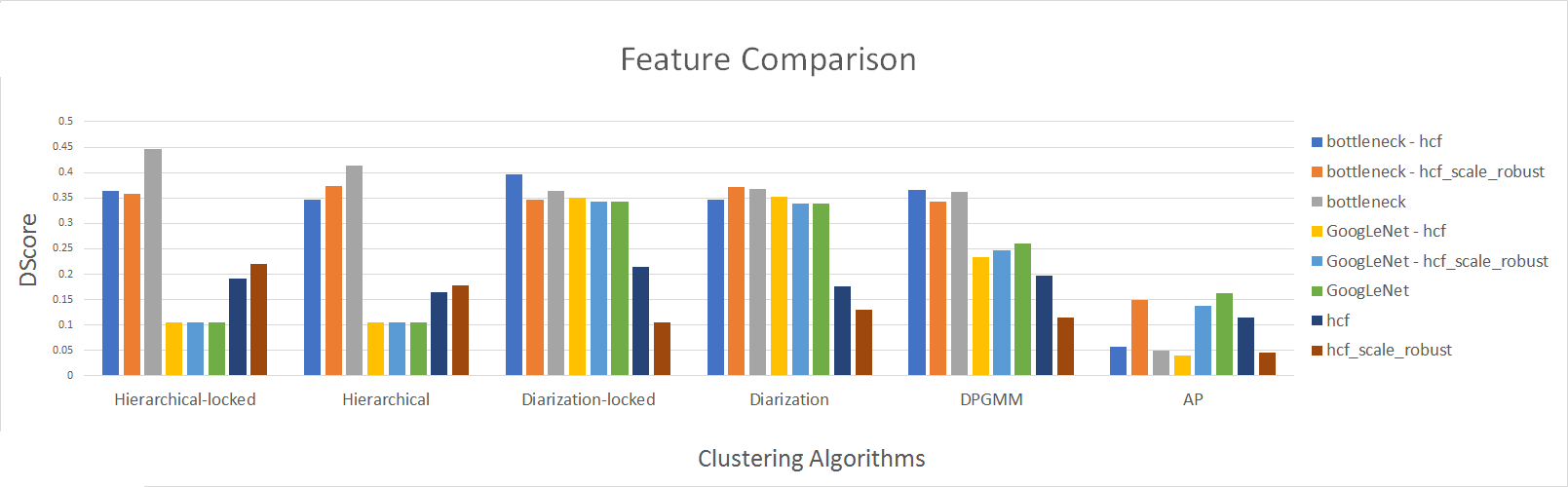}
	\caption{DScore comparison of all the feature types examined. Best viewed in colour}
	\label{fig:results:feat_compare}
\end{figure*}
\begin{figure*}[t]
	\centering
	\includegraphics[width=0.7\linewidth]{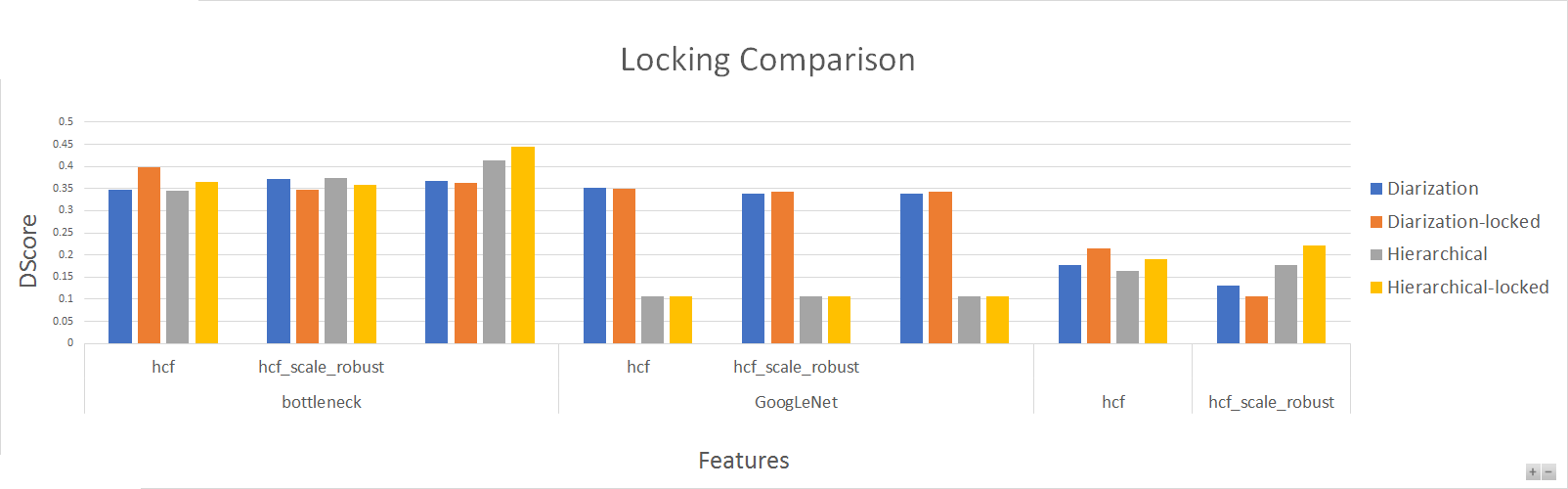}
	\caption{DScore comparison of diarization and hierarchical clustering against their new locking algorithm versions.}
	\label{fig:results:lock_compare}
\end{figure*}

When examining the best results of each clustering algorithm as shown in Table~\ref{tbl:results:clust_compare} it becomes clear that the best performing clustering algorithm was our hierarchical-locked algorithm.
DPGMM and AP clustering were unable to exceed the hierarchical-based clustering algorithms, achieving DScores of 0.37 and 0.16 respectively.
%It should be noted that AP would have very pure clusters but was always oversegmenting, having 44 clusters in the example shown in Fig.~\ref{fig:results:feat_compare}.
Diarization-locked and hierarchical-locked clustering exceeded the performance of their original algorithms, achieving DScores of 0.40 and 0.44 respectively in comparison to their original algorithm DScores of 0.37 and 0.41 respectively.
This is reinforced when we look at the locking comparison graph of Fig.~\ref{fig:results:lock_compare}.
In most cases, particularly when using bottleneck features merging with hcf and when using only hcfs, we can see the benefit gained in using the locking algorithm for diarization and hierarchical clustering.

Also, it was shown that using bottleneck features provided equivalent or improved results for all diarization, hierarchical and DPGMM clustering experiments over that of GoogLeNet features as shown in Fig.~\ref{fig:results:feat_compare} with DScore improvements of up to 0.34.
Hand-crafted features, were never able to outperform the best DCNN features for each experiment achieving DScores of up to 0.21 and 0.22 for the original and scale-robust HCFs respectively.
No great increase was achieved when concatenating DCNN features with hand-crafted features and was, in fact, often detrimental to clustering accuracy.
Overall, the highest DScores were achieved using our improved hierarchical-locked clustering algorithm in combination with DCNN bottleneck features.

%To further prove the benefit of utilizing our initialization and image locking scheme, we calculated the average improvement attained over all hierarchical and diarization clustering tests.
%While not all tests gave an increase in DScore using our technique, there was an average increase of 0.03 across all 40 diarization and hierarchical clustering tests performed.
%The largest increase in DScore due to our intialization method was 0.16 and the largest decrease was 0.06.
%It should be noted that of the six instances where a decrease was recorded, only two were of any significance (i.e. greater than 0.005).
%From this, we concluded that initializing images assumed to be of the same group together, as well as ensuring that those images stay locked in the same cluster, proved beneficial overall.

\subsection{Clustering Qualitative Analysis}\label{sec:results:qual}
\begin{figure}
	\centering
	\includegraphics[width=0.7\linewidth]{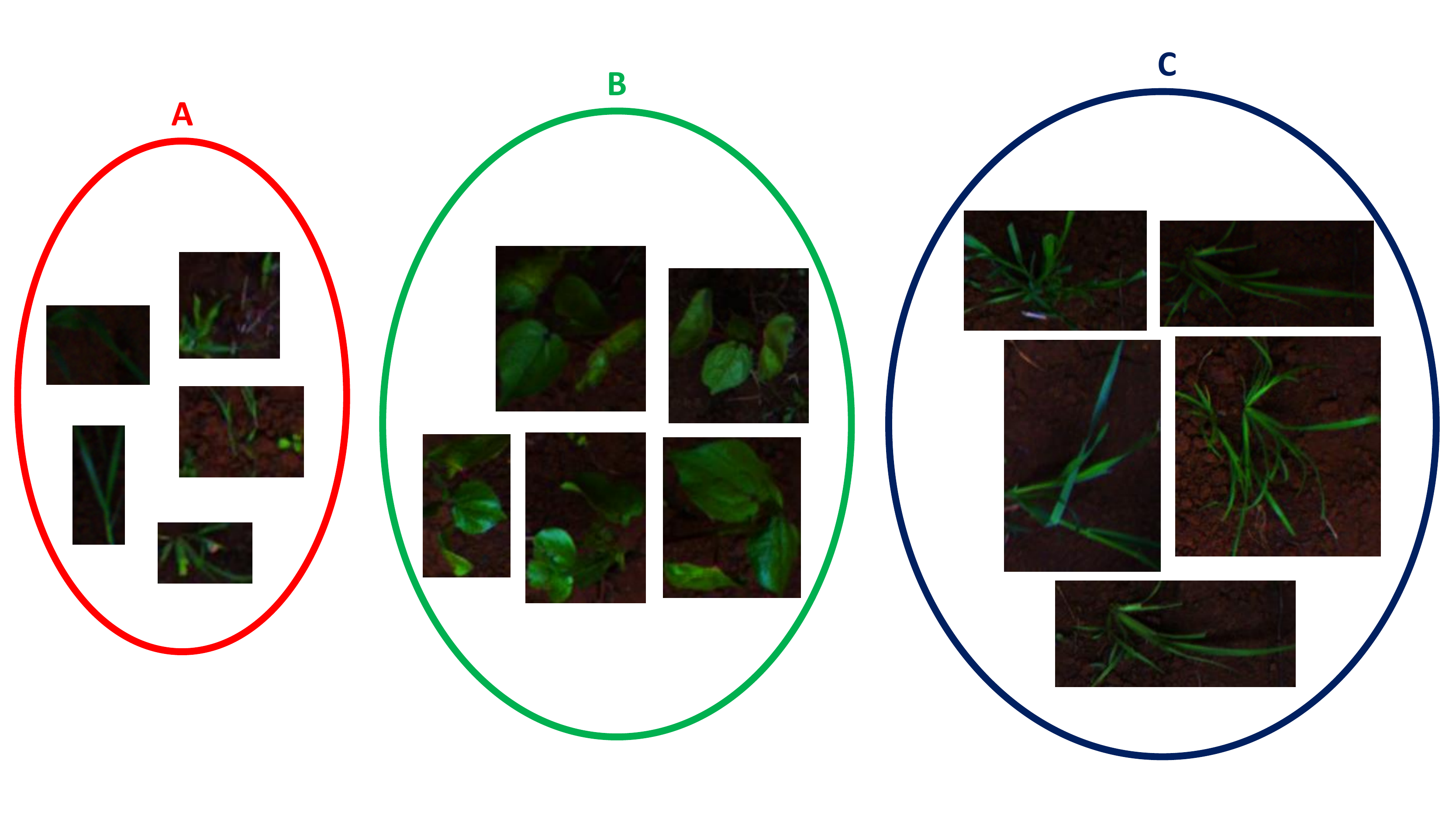}
	\caption{Summary of clustering results for best clustering system (hierarchical-locked clustering using bottleneck features).}
%	 Each cluster group shows the top 5 plant images which are closest to the mean of the cluster. Images have been enhanced for visualization purposes. Best viewed in colour.}
	\label{fig:results:clust_summary}
\end{figure}

As well as evaluating how well each method works in a quantitative manner, we also evaluated the final clustering map  generated in a qualitative manner.
Here we looked at the actual plant images in each cluster, particularly for our most successful hierarchical-locked clustering system.
When we look at the three clusters generated by this technique as summarized in Fig.~\ref{fig:results:clust_summary} we can see that cotton plants appear to have been grouped together in cluster B while the grasses have been distributed into clusters A and C.
Our algorithm was found therefore to almost always group cotton plants together in the same group without any prior knowledge about the appearance of cotton plants.

Through this example we can also see the main source of error in the clustering assignments.
There is no clear distinction made between the two types of grass.
This is to be expected as they are both very visually similar even to human eyes and they could be grouped together by the farmer if desirable when applied to real fields.
As well as this, the sowthistle plants have been grouped together with the grasses as opposed to with the cotton or in its own cluster.
We have attributed this to the fact that the sowthistle plants are particularly young here and have few distinguishing characteristics combined with the fact that there are very few samples of sowthistle, making it more difficult to build a distinctive plant model.
%that there are so few examples of sowthistle with which to build a viable plant model, and that it was not grouped with its fellow broadleaf (cotton) as the cotton cluster likely had a very good model describing what cotton should look like, disqualifying sowthistle from contention.
%\TODO{check that last sentence and my hypothises to check they seem valid.}

\begin{figure}[t]
	\centering
	\includegraphics[width=\linewidth]{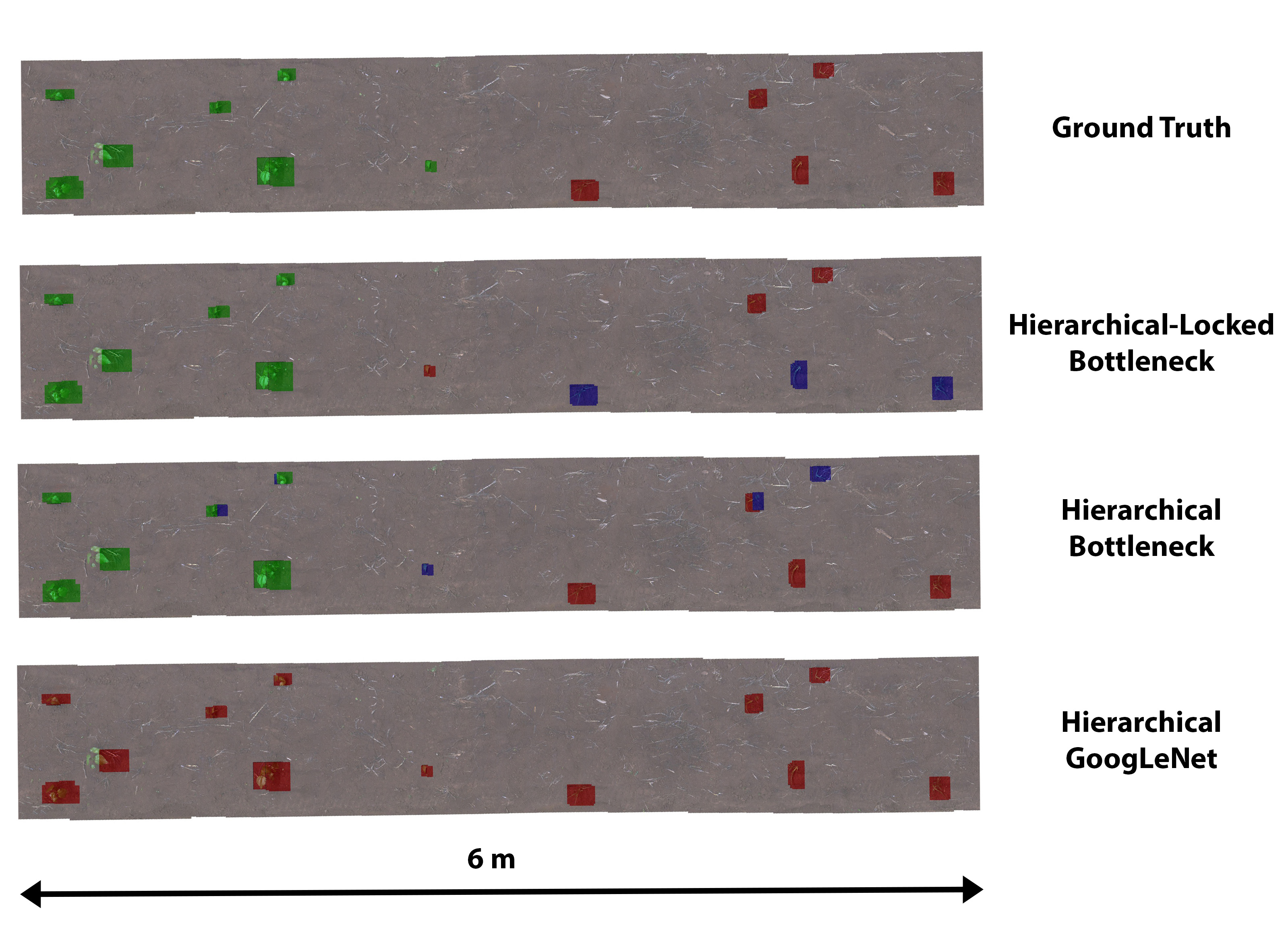}
	\caption{Comparisons of small sections of the final map. Here, colours represent different groups for the clustering algorithms. The main cotton cluster for each has been kept as green. When two colours appear in the same region for Hierarchical with bottleneck features, it is representative of differing assignments being given to different images same plant. Best viewed in colour.}
	\label{fig:results:map}
\end{figure}

Finally in our qualitative analysis, we review the final map as it would be seen by a farmer shown in Fig.~\ref{fig:results:map}.
Here we only show a small region of the field, however an example of a fully mapped field is shown in the supplemental video supplied\footnote[1]{\url{http://tinyurl.com/WS-AGR-QUT}}. % \TODO{do I need a link here?}.
Here it is shown not only what is possible at present (fairly accurate cotton vs grass clustering) but also how much of an improvement can be gained through our hierarchical-locked method using bottleneck features over other techniques which sometimes give the same plant a different clustering assignment when the robot is given multiple images of the same plant.

\section{Conclusions}
This paper demonstrates the first steps towards generating a completely unsupervised weed scouting system for use in agricultural robotics.
Using data collected in the field from an agricultural robotics platform, we cluster plants together into groups using several state-of-the art clustering algorithms.
We utilized low dimensional DCNN bottleneck features, which are proven to outperform both the high-dimensional DCNN features from the GoogLeNet framework, and hand-crafted plant classification features for the task of clustering.
We showed how clustering accuracy could further be improved for hierarchical-type clustering algorithms through locking images identified as the being the same individual plant together at initialization and through the clustering procedures themselves.
We have also identified some of the key challenges to be overcome for future work to achieve results closer to species-specific clustering such as high interclass similarity and low numbers of species samples.
Despite these challenges, we were able to successfully cluster the vast majority of cotton plants into a single, pure species group, separating it from them from any grasses present in the field without any prior training or knowledge about the field.

\section*{Acknowledgments}
We would like to acknowledge and thank the Grains Research and Development Corporation for contributing funds towards this research.

{\small
	\bibliographystyle{IEEEtran}
	\bibliography{refs}

\begin{thebibliography}{10}
\providecommand{\url}[1]{#1}
\csname url@rmstyle\endcsname
\providecommand{\newblock}{\relax}
\providecommand{\bibinfo}[2]{#2}
\providecommand\BIBentrySTDinterwordspacing{\spaceskip=0pt\relax}
\providecommand\BIBentryALTinterwordstretchfactor{4}
\providecommand\BIBentryALTinterwordspacing{\spaceskip=\fontdimen2\font plus
\BIBentryALTinterwordstretchfactor\fontdimen3\font minus
  \fontdimen4\font\relax}
\providecommand\BIBforeignlanguage[2]{{%
\expandafter\ifx\csname l@#1\endcsname\relax
\typeout{** WARNING: IEEEtran.bst: No hyphenation pattern has been}%
\typeout{** loaded for the language `#1'. Using the pattern for}%
\typeout{** the default language instead.}%
\else
\language=\csname l@#1\endcsname
\fi
#2}}

\bibitem{Charles2011}
G.~Charles and T.~Leven, ``Integrated weed management (iwm) for australian
  cotton,'' \emph{Cotton Pest Management Guide}, pp. 88--119, 2011.

\bibitem{Gilbert2013}
N.~Gilbert, ``A hard look at {GM} crops,'' \emph{Nature}, vol. 497, no. 7447,
  pp. 24--26, 2013.

\bibitem{grdc2016}
\BIBentryALTinterwordspacing
GRDC. (2016) Integrated weed management hub. Accessed 07-09-2016. [Online].
  Available: \url{www.grdc.com.au/Resources/IWMhub}
\BIBentrySTDinterwordspacing

\bibitem{Pedersen2006}
S.~M. Pedersen, S.~Fountas, H.~Have, and B.~S. Blackmore, ``Agricultural
  robots—system analysis and economic feasibility,'' \emph{Precision
  Agriculture}, vol.~7, no.~4, pp. 295--308, July 2006.

\bibitem{Sansao2012}
J.~P. Sansao, M.~S. Júnior, L.~A. Mozelli, F.~A. Pinto, and D.~M. Queiroz,
  ``Weed {Mapping} {Using} {Digital} {Images},'' in \emph{Proceedings of
  {International} {Conference} on {Agricultural} {Engineering}-{CIGR}}, 2012.

\bibitem{gerhards2006practical}
R.~Gerhards and H.~Oebel, ``Practical experiences with a system for
  site-specific weed control in arable crops using real-time image analysis and
  gps-controlled patch spraying,'' \emph{Weed Research}, vol.~46, no.~3, pp.
  185--193, 2006.

\bibitem{Grira2004}
N.~Grira, M.~Crucianu, and N.~Boujemaa, ``Unsupervised and semi-supervised
  clustering: a brief survey,'' \emph{A review of machine learning techniques
  for processing multimedia content, Report of the MUSCLE European Network of
  Excellence (FP6)}, 2004.

\bibitem{MacQueen1967}
J.~MacQueen and {others}, ``Some methods for classification and analysis of
  multivariate observations,'' in \emph{Proceedings of the fifth {Berkeley}
  symposium on mathematical statistics and probability}, vol.~1.\hskip 1em plus
  0.5em minus 0.4em\relax Oakland, CA, USA., 1967, pp. 281--297.

\bibitem{Johnson1967}
S.~C. Johnson, ``Hierarchical clustering schemes,'' \emph{Psychometrika},
  vol.~32, no.~3, pp. 241--254, Sept. 1967.

\bibitem{Salvador2004}
S.~Salvador and P.~Chan, ``Determining the number of clusters/segments in
  hierarchical clustering/segmentation algorithms,'' in \emph{16th {IEEE}
  {International} {Conference} on {Tools} with {Artificial} {Intelligence},
  2004. {ICTAI} 2004}, Nov. 2004, pp. 576--584.

\bibitem{Frigui1999}
H.~Frigui and R.~Krishnapuram, ``A robust competitive clustering algorithm with
  applications in computer vision,'' \emph{Pattern Analysis and Machine
  Intelligence, IEEE Transactions on}, vol.~21, no.~5, pp. 450--465, 1999.

\bibitem{Frigui1997}
H.~Frigui and R.~Krishnapuram, ``Clustering by competitive agglomeration,''
  \emph{Pattern recognition}, vol.~30, no.~7, pp. 1109--1119, 1997.

\bibitem{Steinberg2010}
D.~M. Steinberg, S.~B. Williams, O.~Pizarro, and M.~V. Jakuba, ``Towards
  autonomous habitat classification using {Gaussian} {Mixture} {Models},'' in
  \emph{2010 {IEEE}/{RSJ} {International} {Conference} on {Intelligent}
  {Robots} and {Systems} ({IROS})}, Oct. 2010, pp. 4424--4431.

\bibitem{Frey2006}
B.~J. Frey and D.~Dueck, ``Mixture modeling by affinity propagation,''
  \emph{Advances in neural information processing systems}, vol.~18, p. 379,
  2006, 00087.

\bibitem{Dueck2007}
D.~Dueck and B.~J. Frey, ``Non-metric affinity propagation for unsupervised
  image categorization,'' in \emph{2007 {IEEE} 11th {International}
  {Conference} on {Computer} {Vision}}.\hskip 1em plus 0.5em minus 0.4em\relax
  IEEE, 2007, pp. 1--8, 00165.

\bibitem{Wooters2008}
C.~Wooters and M.~Huijbregts, ``The {ICSI} {RT}07s {Speaker} {Diarization}
  {System},'' in \emph{Multimodal {Technologies} for {Perception} of {Humans}},
  ser. Lecture {Notes} in {Computer} {Science}, R.~Stiefelhagen, R.~Bowers, and
  J.~Fiscus, Eds.\hskip 1em plus 0.5em minus 0.4em\relax Springer Berlin
  Heidelberg, 2008, pp. 509--519.

\bibitem{AngueraMiro2012}
X.~Anguera~Miro, S.~Bozonnet, N.~Evans, C.~Fredouille, G.~Friedland, and
  O.~Vinyals, ``Speaker {Diarization}: {A} {Review} of {Recent} {Research},''
  \emph{IEEE Transactions on Audio, Speech, and Language Processing}, vol.~20,
  no.~2, pp. 356--370, Feb. 2012.

\bibitem{Donahue2013}
J.~Donahue, Y.~Jia, O.~Vinyals, J.~Hoffman, N.~Zhang, E.~Tzeng, and T.~Darrell,
  ``Decaf: A deep convolutional activation feature for generic visual
  recognition,'' \emph{arXiv preprint arXiv:1310.1531}, 2013.

\bibitem{Krizhevsky2012}
A.~Krizhevsky, I.~Sutskever, and G.~E. Hinton, ``Imagenet classification with
  deep convolutional neural networks.'' in \emph{NIPS}, vol.~1, no.~2, 2012,
  p.~4.

\bibitem{Szegedy2015}
C.~Szegedy, W.~Liu, Y.~Jia, P.~Sermanet, S.~Reed, D.~Anguelov, D.~Erhan,
  V.~Vanhoucke, and A.~Rabinovich, ``Going {Deeper} {With} {Convolutions},'' in
  \emph{Proceedings of the {IEEE} {Conference} on {Computer} {Vision} and
  {Pattern} {Recognition}}, 2015, pp. 1--9.

\bibitem{philipp2002improving}
I.~Philipp and T.~Rath, ``Improving plant discrimination in image processing by
  use of different colour space transformations,'' \emph{Computers and
  electronics in agriculture}, vol.~35, no.~1, pp. 1--15, 2002.

\bibitem{Suzuki85_1}
S.~Suzuki and K.~Abe, ``Topological structural analysis of digitized binary
  images by border following,'' in \emph{CVGIP}, 1985, pp. 32--46.

\bibitem{hall2015evaluation}
D.~Hall, C.~McCool, F.~Dayoub, N.~S\"underhauf, and B.~Upcroft, ``Evaluation of
  features for leaf classification in challenging conditions,'' in \emph{Winter
  Conference on the Applications of Computer Vision}, 2015.

\bibitem{Haug2014}
S.~Haug, A.~Michaels, P.~Biber, and J.~Ostermann, ``Plant classification system
  for crop/weed discrimination without segmentation,'' in \emph{IEEE Winter
  Conference on Applications of Computer Vision}, 2014.

\bibitem{Bonnet2015}
P.~Bonnet, W.-P. Vellinga, R.~Planqué, A.~Rauber, S.~Palazzo, B.~Fisher, and
  H.~Müller, ``{LifeCLEF} 2015: {Multimedia} {Life} {Species} {Identification}
  {Challenges},'' in \emph{Experimental {IR} {Meets} {Multilinguality},
  {Multimodality}, and {Interaction}: 6th {International} {Conference} of the
  {CLEF} {Association}, {CLEF}'15, {Toulouse}, {France}, {September} 8-11,
  2015, {Proceedings}}, vol. 9283.\hskip 1em plus 0.5em minus 0.4em\relax
  Springer, 2015, p. 462.

\bibitem{ge2015content}
Z.~Ge, C.~Mccool, and P.~Corke, ``Content specific feature learning for
  fine-grained plant classification,'' in \emph{Working notes of CLEF 2015
  conference}, 2015.

\bibitem{chen1998speaker}
S.~Chen and P.~Gopalakrishnan, ``Speaker, environment and channel change
  detection and clustering via the bayesian information criterion,'' in
  \emph{Proc. DARPA Broadcast News Transcription and Understanding
  Workshop}.\hskip 1em plus 0.5em minus 0.4em\relax Virginia, USA, 1998, p.~8.

\bibitem{Zhou2000}
B.~Zhou and J.~Hansen, ``Unsupervised {Audio} {Stream} {Segmentation} {And}
  {Clustering} {Via} {The} {Bayesian} {Information} {Criterion},'' in \emph{in
  {Proc}. {ISCLP} 2000}, 2000, pp. 714--717.

\bibitem{Schwarz1978}
G.~Schwarz and {others}, ``Estimating the dimension of a model,'' \emph{The
  annals of statistics}, vol.~6, no.~2, pp. 461--464, 1978.

\bibitem{Ferguson1973}
T.~S. Ferguson, ``A {Bayesian} analysis of some nonparametric problems,''
  \emph{The annals of statistics}, pp. 209--230, 1973, 03840.

\bibitem{Rasmussen1999}
C.~E. Rasmussen, ``The infinite {Gaussian} mixture model.'' in \emph{{NIPS}},
  vol.~12, 1999, pp. 554--560.

\bibitem{Rublee2011}
E.~Rublee, V.~Rabaud, K.~Konolige, and G.~Bradski, ``{ORB}: {An} efficient
  alternative to {SIFT} or {SURF},'' in \emph{2011 {International} {Conference}
  on {Computer} {Vision}}, Nov. 2011, pp. 2564--2571.

\bibitem{Ruckelshausen2009}
A.~Ruckelshausen, P.~Biber, M.~Dorna, H.~Gremmes, R.~Klose, A.~Linz, F.~Rahe,
  R.~Resch, M.~Thiel, D.~Trautz, and {others}, ``{BoniRob}–an autonomous
  field robot platform for individual plant phenotyping,'' \emph{Precision
  agriculture}, vol.~9, no. 841, p.~1, 2009.

\bibitem{Manning2008}
C.~D. Manning, P.~Raghavan, and {others}, \emph{Introduction to information
  retrieval}, 2008, vol.~1.

\end{thebibliography}
}

\end{document}